\documentclass[10pt,twocolumn,letterpaper]{article}

\usepackage{cvpr}              
\usepackage{graphicx}
\usepackage[ruled,vlined,linesnumbered,commentsnumbered]{algorithm2e}
\usepackage{amsmath} 
\usepackage{bm} 
\usepackage{multirow}
\usepackage[table]{xcolor}
\usepackage[utf8]{inputenc}
\SetKwComment{Comment}{$\triangleright$\ }{} 

\usepackage{multibib}
\newcites{supp}{Supplementary References}

\definecolor{lightpink}{RGB}{233, 23, 115}
\definecolor{first}{RGB}{220, 245, 220}   
\definecolor{second}{RGB}{255, 255, 200}  
\definecolor{third}{RGB}{200, 230, 255}   

\definecolor{cvprblue}{rgb}{0.21,0.49,0.74}
\usepackage[pagebackref,breaklinks,colorlinks,allcolors=cvprblue]{hyperref}

\def\paperID{***} 
\def\confName{CVPR}
\def\confYear{2026}

\title{EARTalking: End-to-end GPT-style Autoregressive Talking Head Synthesis with Frame-wise Control}

\author{Yuzhe Weng$^1$, Haotian Wang$^1$, Yuanhong Yu$^3$, Jun Du$^{1 \ast}$, \\ \vspace{0em}
Shan He$^2$, Xiaoyan Wu$^2$, Haoran Xu$^2$ \\ \vspace{0.2em}
$^1$ University of Science and Technology of China,  $^2$ iFLYTEK, $^3$ Zhejiang University \\
\footnotesize\textsuperscript{*}Corresponding author: jundu@ustc.edu.cn
\vspace{-2.0em}
}

\begin{document}
\twocolumn[{
\renewcommand\twocolumn[1][]{#1}%
\maketitle

\begin{center}
    \centering
    \captionsetup{type=figure}
    \includegraphics[width=1.0\linewidth]{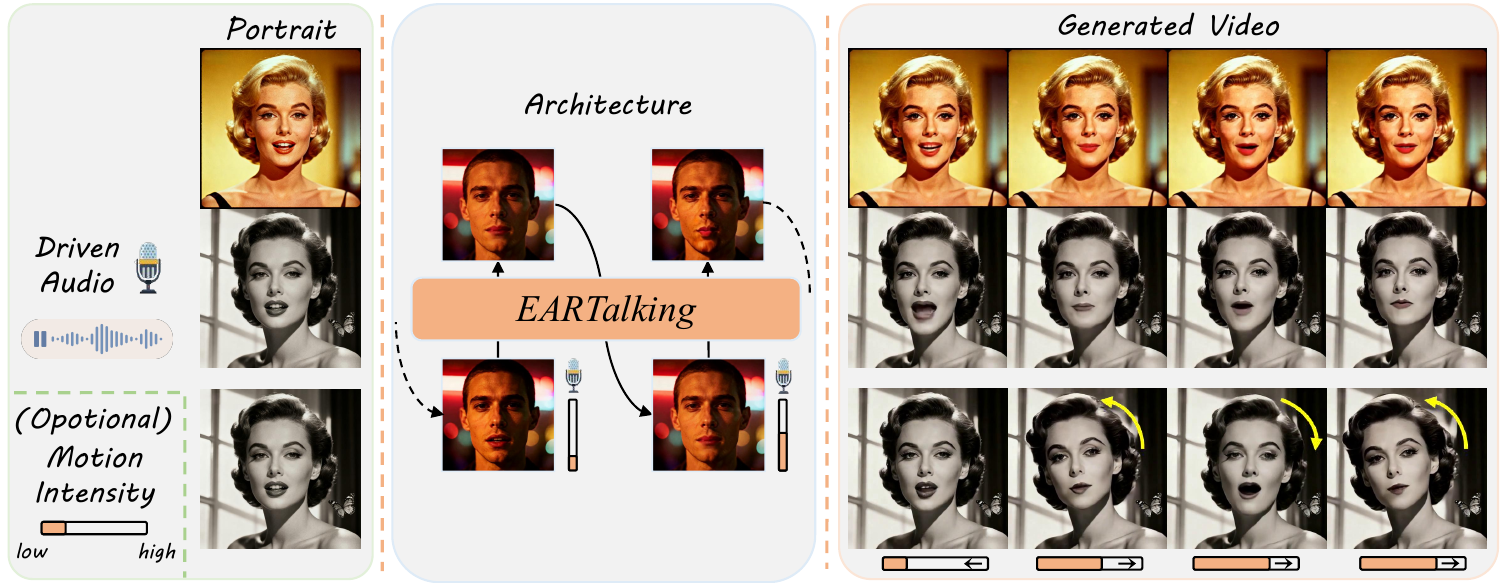}
    \caption{The autoregressive generation paradigm and results. EARTalking proposes a novel effective architecture for the Talking Head Generation task, supporting streaming control and streaming generation through a more concise and effective context learning approach.}
\end{center}
}]

\begin{abstract}

Audio-driven talking head generation aims to create vivid and realistic videos from a static portrait and speech. Existing AR-based methods rely on intermediate facial representations, which limit their expressiveness and realism. Meanwhile, diffusion-based methods generate clip-by-clip, lacking fine-grained control and causing inherent latency due to overall denoising across the window.
To address these limitations, we propose \textbf{EARTalking}, the novel \textbf{end-to-end, GPT-style autoregressive model} for interactive audio-driven talking head generation. Our method introduces a novel frame-by-frame, in-context, audio-driven streaming generation paradigm. 
For inherently supporting variable-length video generation with identity consistency, we propose the Sink Frame Window Attention (SFA) mechanism.
Furthermore, to avoid the complex, separate networks that prior works required for diverse control signals, we propose a streaming Frame Condition In-Context (FCIC) scheme. This scheme efficiently injects diverse control signals in a streaming, in-context manner, enabling interactive control at every frame and at arbitrary moments.
Experiments demonstrate that EARTalking outperforms existing autoregressive methods and achieves performance comparable to diffusion-based methods. 
Our work demonstrates the feasibility of in-context streaming autoregressive control, unlocking a scalable direction for flexible, efficient generation. The code will be released for the reproducibility.

\end{abstract}

\section{Introduction}
\label{sec:intro}
Talking head generation (THG), the task of synthesizing realistic videos from static images and audio, is essential for applications such as virtual avatars, film dubbing, and personalized content creation, and thus has garnered significant research interest\cite{thdsurvey}. Although recent diffusion-based methods\cite{xu2024hallo, tian2024emo, chen2025echomimic} show remarkable progress, they face critical challenges. Functionally, diffusion-based methods introduce inherent algorithmic latency because the denoising steps must be applied to latent variables across an entire time window. In other words, the model must process the entire window as a single batch before any frame, including the first, can be output. In terms of control, effective THG must process both speech (for lip-sync) and para-linguistic information for naturalness. However, previous methods apply control globally, which lacks fine-grained temporal control\cite{Gan_2023_ICCV, EmotiveTalk} or rely on complex auxiliary networks\cite{PC-Talk}, which severely hinders the scalability of the model. Furthermore, previous work often relied on unidirectional attention (from vision to audio), which hindered the model's ability to fully fuse audio-visual information. Efficient bidirectional fusion is essential for achieving accurate lip synchronization and natural expression. 

Grounded in the intrinsically streaming features of the audio modality, we seek to develop a generation pipeline for streaming-input-driven talking heads that offers greater controllability and extensibility. Recent studies have explored non-end-to-end (non-E2E) AR approaches\cite{chu2025artalk, sungbin24_interspeech}, which typically rely on a multi-stage process: first predicting intermediate representations (e.g., facial motion), and then rendering these representations into video. The non-E2E design fundamentally hinders the model's ability to capture and generate fine-grained details at both the pixel and semantic levels, which severely degrades the fidelity and lip-sync accuracy of the resulting video, also increases model complexity and presents significant scaling challenges.

To address these challenges, we introduce EARTalking, a novel end-to-end, frame-by-frame autoregressive model for talking head generation.
We propose the Sink Frame Window Attention (SFA), a novel training and inference mechanism. SFA enables variable-length inference capabilities while being trained exclusively on fixed-length sequences. During inference, the reference image and information from previously generated segments are stored in a kv-cache.  
To maintain consistency between the generated dynamic portraits and the reference image, the reference image is designated as a sink frame in the attention window. This allows it to act as a stable anchor feature for the adaLN conditioning of all subsequent frames.
Furthermore, EARTalking introduces the Frame Condition In-Context (FCIC) method, a novel approach that adapts the in-context learning paradigm for the specific demands of talking head video generation. This method facilitates the injection of cross-modal conditions into the talking head video stream on a per-frame basis with a concise in-context format. 
Consequently, this approach eliminates the need to design specialized modules for each new condition. Instead, new functionalities can be integrated simply by training the model with new control conditions supplied as context to meet different requirements.
Furthermore, FCIC enables interactive, customized control at each time step, crucially without affecting adjacent frames. 

In summary, our main contributions are as follows: 
\begin{itemize}
    \item We introduce \textbf{EARTalking}, the novel GPT-style end-to-end autoregressive framework for frame-by-frame audio-driven talking head generation, which offers a superior framework for streaming generation and control.
    
    \item We propose a \textbf{Sink Frame Window Attention (SFA)} mechanism, which integrates \textbf{adaLN Sink} and \textbf{kv cache}, which effectively mitigates autoregressive error accumulation while ensuring fidelity to the reference image. This design enables variable-length inference based on fixed-length training.
    
    \item Our \textbf{Frame Condition In-Context (FCIC)} scheme enables the streaming input of control conditions via in-context inputs, providing the capability for fine-grained, streaming interactive control at arbitrary temporal frames. This eliminates the need for complex, redundant auxiliary networks for new control modalities. 
\end{itemize}

\section{Related Work}
\label{sec:formatting}

\begin{figure*}[t] 
    \centering
    
    \includegraphics[width=\textwidth, page=1]{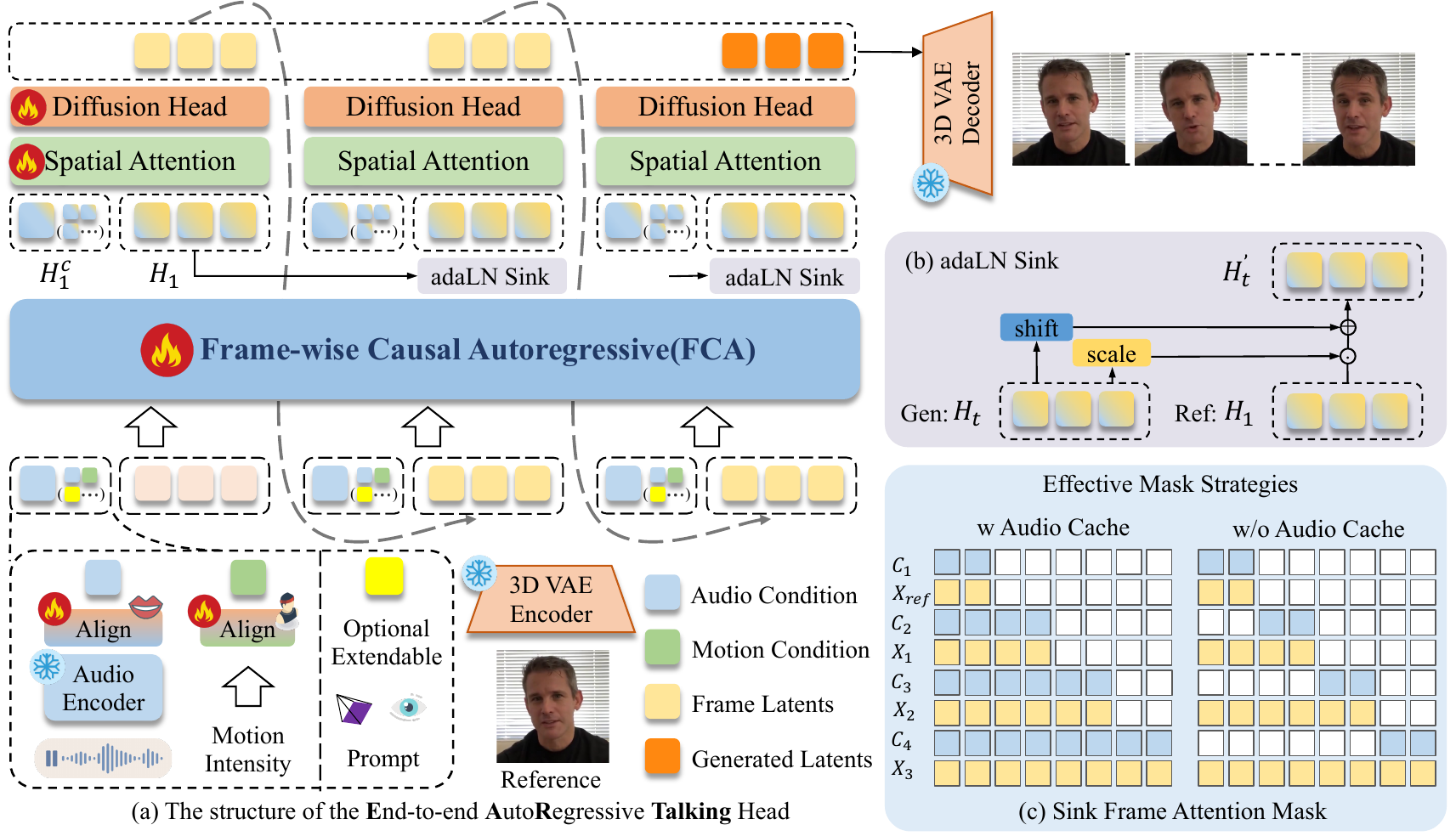}
    
    \caption{The overall framework of EARTalking (a), illustrating the causal autoregressive pipeline for inference and training, which follows the depicted paradigm for frame-by-frame conditional input and generation. (b) The adaLN sink mechanism, which enhances consistency by anchoring generated frames to the reference frame. (c) Two causal attention mask strategies validated as effective during training.} 
    \label{fig: overall framework}
\end{figure*}

\subsection{Video Generation}

Advances in generative diffusion models have significantly accelerated video synthesis. The early U-Net models \cite{SVD} used spatiotemporal attention to maintain temporal consistency.
Inspired by the success of vision transformer\cite{ViT}, the Diffusion Transformer\cite{DiT} design becomes a scalable, high-performance backbone for diffusion models and a mainstream approach in video generation. A dominant technical path\cite{kong2024hunyuanvideo, wan2025} now combines a DiT backbone with 3D RoPE and spatiotemporal compression techniques like 3D VAEs.
Furthermore, autoregressive models, which have been proven strong and scalable in multimodal understanding\cite{xie2024show, chen2025blip3, xie2025muse}, are also showing considerable potential for video generation\cite{wang2024emu3, deng2024nova, yu2025videomar}.

\subsection{Audio-driven Talking Head Video Generation}

Frontier technologies in video generation have seen explosive growth in recent years. As a downstream task, Talking Head Video Generation has also consequently experienced a parallel boom\cite{cheng2024dawn, EmotiveTalk, tan2024flowvqtalker, jang2024faces, peng2024synctalk}.
Early approaches\cite{StyleTalk, ma2023dreamtalk}, predominantly leveraging GANs, often relied on predicting intermediate representations such as facial landmarks or 3D Morphable Models (3DMMs), which were then rendered into video. Subsequent works, such as ARTalk\cite{chu2025artalk} and MultiTalk\cite{sungbin24_interspeech}, adopted autoregressive methods to first generate 3D motion and then synthesize it into video, thereby improving generation speed. 
More recently works\cite{tian2024emo, xu2024hallo, chen2025echomimic} have built upon pre-trained image diffusion models to achieve high-fidelity talking head synthesis. Further advancements, including EchoMimicV3\cite{meng2025echomimicv3}, are now leveraging pre-trained DiT-based video diffusion models to enhance the naturalness and realism of the generated talking heads.

\subsection{Controllable Talking Head Generation}

In the specialized field of talking head video generation, achieving robust multi-conditional control, particularly including audio, has long been an area of significant interest. Common controls include expression and motion. Prior work\cite{nocentini2024emovocaspeechdrivenemotional3d, EmotiveTalk} often relied on discrete codebooks or a single representation to exert coarse-grained control over the entire video's performance. Furthermore, many studies have widely investigated diverse control conditions\cite{ma2025exploring, li2025portrait, chen2025taoavatar} and different control schemes\cite{hu2025HunyuanVideo-Avatar}, opening up new possibilities for talking head generation.

\section{Method}

The overall framework of EARTalking is illustrated in Figure \ref{fig: overall framework}. Section \ref{3.1 Preliminary} will detail the preliminaries and theoretical foundations of this work. Following this, in Section \ref{3.2 Frame Causal Control and Generation}, we will focus on the proposed model architecture. In Section \ref{3.3 Sink Frame Window Attention Framework}, we will highlight the  Sink Frame Window Attention Framework(SFA) scheme and the natural suitability of the adaIN sink for the talking head video generation task. Finally, Section \ref{3.4 Frame Condition In-Context Control} will describe the Frame-Conditioned In-Context (FCIC) control, including the distinct training stages and the inference pipeline.

\subsection{Preliminary}
\label{3.1 Preliminary}
\textbf{Task Definition}
The core task of talking head generation is to synthesize a dynamic video $\hat{X} = \{\hat{X}_t\}_{t=1}^T$ given two primary inputs: a static portrait image $I_{\mathrm{ref}}$ and an audio sequence $C= \{C_t\}_{t=1}^{T_a}$. The synthesized video $\hat{X}$ must maintain identity consistency with $I_{\mathrm{ref}}$, achieve accurate lip synchronization with $C$, and exhibit natural expressions and realistic head poses.


\noindent\textbf{Diffusion Loss for Autoregressive Models}
\label{diffloss}
Autoregressive language models typically utilize a cross-entropy loss, which is inherently suitable for the discrete nature of natural language data.
For visual data, autoregressive generation operates within a continuous latent space encoded by a 3D VAE. Quantizing these continuous latents to utilize a discrete cross-entropy loss introduces quantization errors, which degrade the final generation quality.

To address this,  Diffusion Loss\cite{mar} can be used to supervise the continuous probability distribution $p(x|z)$ directly. The loss is formulated as:
\begin{equation}
\mathcal{L}(z, x) = \mathbb{E}_{\epsilon, t} \left[ \left\| \epsilon - \epsilon_\theta(x_t | t, z) \right\|^2 \right].
\end{equation}
Here, $z$ represents the conditional hidden latent, $x$ is the ground-truth continuous latent token, $t$ is a randomly sampled noise timestep, and $\epsilon \sim \mathcal{N}(0, \mathbf{I})$ is a random noise vector. The term $x_t$ (defined as $x_t = \sqrt{\bar{\alpha}_t}x + \sqrt{1-\bar{\alpha}_t}\epsilon$) is the noised version of $x$ at timestep $t$. $\epsilon_\theta(x_t | t, z)$ is the noise-prediction network, parameterized by $\theta$, which is conditioned on $z$ and $t$ to predict the original noise $\epsilon$ from $x_t$. This loss function trains the model by minimizing the mean squared error between the actual noise added ($\epsilon$) and the noise predicted by the network ($\epsilon_\theta$).

\subsection{Frame Causal Control and Generation}
\label{3.2 Frame Causal Control and Generation}

EARTalking is an A2V-AR backbone, which aims to reframe the Talking Head Video Generation task in an LLM-like manner of frame-by-frame understanding and generation In this paradigm, the video generation model can receive distinct multimodal contexts at each frame to dynamically control the generation of subsequent video frames.

\noindent\textbf{Frame-wise Causal Autoregression}
We adopt the $3\text{D}$ Variational Autoencoder ($3\text{D VAE}$) from Wan 2.1\cite{wan2025} to perform $4\times$ temporal and $8\times8$ spatial compression. 
In the Frame-wise Causal Autoregression model (FCA), the generation of each frame is conditioned on the current timestep's audio embedding and the previously generated frames, which serve as inputs. We implement a causal mask during training. This mask is designed to 1) allow the current timestep's audio and visual embeddings to mutually attend to one another, and 2) permit the model to attend to all previously generated frames and all audio embeddings up to the current timestep.

The objective of this frame-by-frame autoregressive generative model is to optimize the joint probability distribution $p(C_1, \dots, C_T, X_{\mathrm{ref}}, X_1, \dots, X_T)$, which can be factorized autoregressively as:

\begin{equation}
p(\mathbf{C}, X_{\mathrm{ref}}, \mathbf{X}) = \prod_{t=1}^{T} p(X_t | \mathbf{C}_{\le t}, X_{\mathrm{ref}}, \mathbf{X}_{\le t-1})
\end{equation}

To simplify notation, we denote the complete sequences as $\mathbf{C} = \{C_t\}_{t=1}^T$ and $\mathbf{X} = \{X_t\}_{t=1}^T$. We further define the history notations $\mathbf{C}_{\le t} = \{C_i\}_{i=1}^t$ to represent the sequence of $C$ up to and including timestep $t$, and $\mathbf{X}_{\le t-1} = \{X_i\}_{i=1}^{t-1}$ to represent the sequence of $X$ before timestep $t$.

Here, $C_t = \{c_\mathrm{a}, c_\mathrm{m}, \dots\}$ represents the set of condition features (e.g., audio, motion) at time $t$ that drive the talking head's movement, $X_{\mathrm{ref}}$ is the set of representation features for the reference image, and $X_t$ denotes the set of representation features for the generated visual frame at time $t$. Each generated frame $X_t$ is produced conditioned on all features up to the current time, the reference frame features $X_{\mathrm{ref}}$, and all previously generated frame features. This in-context-controlled autoregressive paradigm facilitates mutual attention between the audio and visual modalities.

\noindent\textbf{Spatial Masked Autoregression}
At timestep $t$, the Frame-wise Causal Autoregression (FCA) model, leveraging temporal causal attention, outputs two feature sets: $H_t$, which encapsulates coherent motion and detail for the current frame based on past video and speech data, and $H_t^c$, which represents self-attended speech features further aligned with the visual content. Concurrently, a separate feature, $H'_t$, is derived by normalizing $H_1$ and $H_t$ using an adaLN sink network, a process we will detail in Section \ref{3.3 Sink Frame Window Attention Framework}. The primary objective of this section, Spatial Masked Autoregression, is to generate the current frame spatially, conditioned on both $H'_t$ and $H_t^c$. 

The backbone adopts the Masked Autoregressive Generative structure\cite{mar}, a notable approach in the field of image generation. Specifically, we randomly partition the complete set of features representing a single frame into $K$ disjoint groups. The generation process proceeds iteratively in $K$ steps. In each step $k$, the model autoregressively generates a new set of tokens, $Z_t^k$, conditioned on the given conditions $H'_t$ and $H_t^c$ and the $k-1$ token groups generated in previous steps. Finally, these $K$ groups are concatenated to form the complete token sequence representing the entire image. The joint probability distribution of this process is formulated as:

\begin{equation}
p(H_{t}^c, H'_t, \mathbf{Z}_t) = \prod_{k=1}^{K} p(Z_t^k | H_{t}^c, H'_t, \mathbf{Z}_t^{\le k-1})
\end{equation}

During this inference process, $H_{t}^c$ further guides the generation of motion details for the current frame, ensuring they align with the speech condition. Concurrently, $H'_t$ guides the frame's image details and resulting motion, guaranteeing both high-fidelity details and natural movement in the generated frame.

where $Z_t^k$ represents the subset of tokens generated at step or group $k$. For all generated tokens, we train the model using the supervision scheme described in Section \ref{diffloss}. At inference time, the model employs a standard denoising procedure\cite{mar} to sample and denoise based on latent $z$, thereby obtaining $x$.

\subsection{Sink Frame Window Attention Framework}
\label{3.3 Sink Frame Window Attention Framework}

Our backbone network uses the reference image's features as an anchor. The features of subsequently generated frames are mapped onto this anchor via adaLN, which guarantees the model's faithfulness to the reference image during synthesis. Instead of relying on variable-length training or clip-overlapping schemes (which compromise control flexibility) to achieve variable-length inference, our proposed Sink Frame Window Attention method achieves natural, variable-length inference while being trained at a fixed length. The detailed inference process is shown in Algorithm \ref{alg:frame_wise_inference}.

We adopted the adaLN sink solution to maintain better athletic performance and consistency. This method first processes the reference image features with a Frame-wise Causal Autoregression (FCA) transformer to create the sink anchor features $H'_1$. For each subsequent frame, the FCA transformer generates corresponding features $H_t$. These $H_t$ features are then fed into an AdaLayerNorm network to compute scale and shift values, which are applied to the sink anchor $H'_1$ to produce the new frame's representation. The specific implementation is as follows:
\begin{equation}
H'_{t} = \text{adaLN}(H'_{1}, H_{t}) = \sigma(H_{t}) \left( \frac{H'_{1} - \mu}{\sigma} \right) + \mu(H_{t})
\end{equation}

\begin{figure}[t] 
    \centering
    \includegraphics[width=\columnwidth]{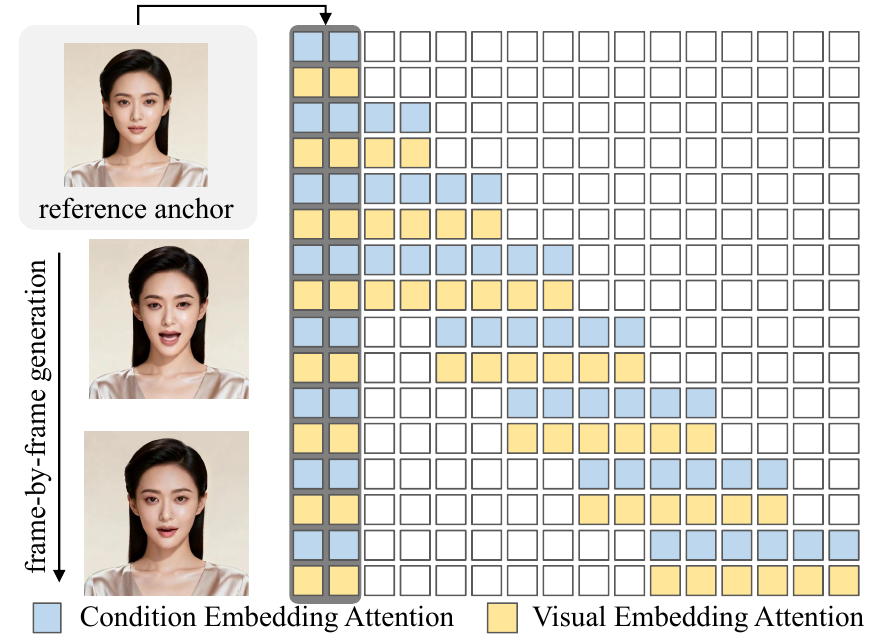}
    \vspace{-15pt}
    \caption{Sink Frame Window Attention}
    \label{fig:sink_adaLN}
    \vspace{-8pt}
\end{figure}

This approach enhances consistency with the reference image and mitigates error accumulation. During training, we perturb input video frames with varying noise intensities to improve the model's self-correction capabilities and robustness.During inference, the Frame-wise Causal Transformer uses a key-value (kv) cache and an attention temporal window ($w$) with a maximum length of $N$ frames (Figure \ref{fig:sink_adaLN}). The kv pairs from the first $N$ frames are sequentially appended to the cache. Starting from the $(N+1)$-th frame, a sliding window mechanism activates: as each new frame is processed, the KV cache corresponding to the second frame in the window is discarded.

Crucially, the kv cache from the first frame, which stores the intermediate hidden states of the reference image, is permanently retained throughout the generation process. This strategy serves two purposes: it ensures that the fine-grained details and identity (ID) of the generated video remain consistent with the reference image, and it significantly mitigates the cumulative errors often introduced by autoregressive models.

To accommodate this specific generation process, we have designed a simple yet effective position embedding scheme\cite{CPE}. Although the prevalent Rotary Position Embedding (RoPE) has demonstrated a certain degree of extrapolation capability, its performance can still degrade during inference when encountering position embeddings unseen during training. Our solution involves maintaining a cyclic temporal absolute position embedding table, denoted as $P=\{p_n\}_{n=1}^{2N-1}$. During inference, the features of the reference image are invariably assigned the first position embedding, $p_1$. Each subsequent frame sequentially adopts the next embedding $p_n$ from the cyclic table. Once the table $P$ is fully traversed, the sequence cycles back, restarting from $p_2$ (as $p_1$ is perpetually reserved for the reference image). The specific sampling rules for position embedding are shown below:

\begin{equation}
p_t = \begin{cases}
p_t & \text{if } t \le 2N-1 \\
p_{(t-1)\%(2N-2) +1}  & \text{if } t > 2N-1
\end{cases}
\end{equation}

During training, the reference image's position embedding is likewise fixed to $p_1$. The subsequent $N-1$ frames are then assigned $N-1$ consecutive position embeddings, sampled starting from a random offset within the cyclic table $P$. Given that the maximum sliding window length is $N$ and the table size is $2N-1$, the model can automatically learn this cyclic ordering during training. This design ensures that no positional ambiguity arises and that the model is exposed to all possible relative positional combinations within the attention window. This training strategy effectively enhances the model's extrapolation capabilities for autoregressive talking head video generation.

\subsection{Frame Condition In-Context Control}
\label{3.4 Frame Condition In-Context Control}

Furthermore, we integrate multi-modal control signals using a frame-by-frame, \emph{in-context} approach. The model implicitly learns the correlations and attentional weights between visual features and auxiliary inputs (such as audio features) directly within its self-attention mechanisms. This architecture facilitates the integration of new control modalities in an elegant and straightforward manner. For the task of talking head generation, we employ distinct frame-level control signals: audio, for which representations are extracted using WavLM-Large\cite{chen2022wavlm}; motion intensity, which is quantified as the frame-wise mean Euclidean distance of facial edge landmarks. Additionally, we employ a motion intensity encoding scheme that is consistent with the absolute positional encoding scheme. This method maps the numerical intensity value into an embedding vector, making its variations easily perceptible to the model. During inference, the generation of the current frame is conditioned on the previously generated frame as well as the current and past control conditions. Our experiments validate the efficacy and strong audio-visual alignment capabilities of our frame-level, in-context, autoregressive control strategy. A key insight from this approach is that the paradigm can be simply and effectively extended to incorporate additional control conditions without necessitating customized or external modifications to the network architecture. This inherent extensibility unlocks significant potential for future advancements in controllable talking head generation.

\begin{algorithm}[h]
\caption{Frame-wise A2V-AR Inference}
\label{alg:frame_wise_inference}
\KwIn{
    Reference portrait $I_{\mathrm{ref}}$; \\
    Driving condition sequence $\mathcal{C} = \{C_t\}_{t=1}^T$; \\
    VAE Encoder $\mathcal{E}$; \\
    DDPM sampler $\mathcal{D}$; \\
}

\KwOut{Sequence of video latents $\mathcal{X} = \{X_t\}_{t=1}^T$}

$X_{\mathrm{ref}} \leftarrow \mathcal{E}(I_{\mathrm{ref}})$\;

$KV_{cache} \leftarrow []$\;
$\mathcal{X} \leftarrow []$\;

\For{$t \leftarrow 1$ \text{to} $T$}{
    
    \If{$t = 1$}{ \Comment{Frame-wise AR network}
        $H_t^c, H_t, KV_{cache} \leftarrow f_\theta(C_1, X_{\mathrm{ref}})$\;
        $H'_{1} \leftarrow H_t$\; 
    }
    \Else{
        $H_t^c, H_{t}, KV_{cache} \leftarrow f_\theta(C_t, X_{t-1}, KV_{cache})$\; 
        $H'_t \leftarrow H'_{1} \odot \text{ScaleProj}(H_{t}) + \text{ShiftProj}(H_{t})$\;
    }
    
    $\mathcal{Z} \leftarrow []$\;
    
    \For{$k \leftarrow 1$ \text{to} $K$}{ \Comment{Masked AR network}
        $Z_k \leftarrow m_\theta(H_t^c, H'_t, Z_{\le k-1})$\; 
        Append $Z_k$ to $\mathcal{Z}$\; 
    }
    
    Sample $\bm{\varepsilon} \sim \mathcal{N}(0, \mathbf{I})$\;
    $X_t \leftarrow \mathcal{D}(\mathcal{Z}, \bm{\varepsilon}, T_{steps})$\;

    Append $X_t$ to $\mathcal{X}$\;
}

\KwRet{$\mathcal{X}$}\;

\end{algorithm}

\section{Experiment}
\subsection{Setup}

\textbf{Implementation Details.}
Our model was trained on the filtered HDTF\cite{hdtf}, MEAD\cite{mead}, and Hallo3\cite{hallo3} datasets. The model weights were initialized from an autoregressive video generation backbone\cite{deng2024nova}. Training utilized $512 \times 512$ resolution videos with a batch size of 4. We employed a two-stage training strategy, using the AdamW optimizer for both stages. In the first stage, the learning rate was set to 1e-4. During this stage, we used the 3D VAE from Wan 2.1\cite{wan2025} to extract visual latents and trained the audio alignment network and the backbone model. In the second stage, the learning rate was reduced to 1e-5. We replaced sinusoidal positional encoding\cite{Transformer} with cyclic temporal positional encoding and introduced motion intensity control. The primary training window was 37 video frames; ablation studies with different window lengths are detailed in the supplementary material. Unless otherwise specified, all reported results are based on an inference configuration where the model operates with a 10-frame latent sliding window, the Spatial Masked Autoregression generation is set to 36 steps, and the diffusion head employs 25 steps for denoising.

\noindent\textbf{Evaluation Metrics.} We assess our model's performance using standard metrics for video and talking head generation: Fr'{e}chet Inception Distance (FID)\cite{fid}, Fr'{e}chet Video Distance (FVD)\cite{fvd}, Expression-FID (E-FID)\cite{tian2024emo}, Synchronization-C (Sync-C), and Synchronization-D (Sync-D)\cite{stylegan_v}. For visual consistency, we use FID$\downarrow$, and for expression consistency, we use E-FID$\downarrow$. Temporal dynamics are evaluated using FVD$\downarrow$. Finally, audio-lip synchronization is quantified by Sync-C$\uparrow$ and Sync-D$\downarrow$. Avg-R$\downarrow$ refers to the average ranking of the model across various metrics, measuring its overall performance.

\begin{table*}[t]
\centering

\resizebox{\textwidth}{!}{
\tiny 
\begin{tabular}{c|c|c|c|c|c|c|c|c|c}
\toprule
\textbf{Dataset} & \textbf{Method}  & \textbf{Params} & \textbf{Avg-R} ($\downarrow$) & \textbf{FID} ($\downarrow$) & \textbf{FVD} ($\downarrow$) & \textbf{Sync-C} ($\uparrow$) & \textbf{Sync-D} ($\downarrow$) & \textbf{E-FID} ($\downarrow$) \\
\midrule
\multirow{6}{*}{HDTF}
& AniPortrait & 2B & 3.50 & \underline{17.629} & 443.902 & 3.368 & 10.712 & 2.210 \\
& EchoMimic & 2B & \underline{2.66} & 18.733 & 629.370 & 5.610 & 8.953 & \textbf{0.803} \\
& EchoMimicV3 & 1.3B & 3.83 & 21.054 & \underline{380.812} & 2.824 & 11.794 & 1.987 \\
& AniTalker & 0.1B & 2.83 & 34.644 & 476.710 & \underline{5.639} & \textbf{8.588} & 1.551 \\
& Ditto & 0.1B & 3.00 & \textbf{16.253} & 384.232 & 4.036 & 10.250 & 2.861 \\
\rowcolor{first}
& Ours & 0.6B & \textbf{1.66} & 18.981 & \textbf{363.909}  & \textbf{5.707}  & \underline{8.903} & \underline{1.326} \\
\midrule
\multirow{6}{*}{MEAD}
& AniPortrait & 2B & 4.16 & 63.460 & 519.822 & 1.324 & 12.650 & 1.886 \\
& EchoMimic & 2B & 3.00 & 51.546 & 775.951 & 5.276 & 9.565 & 1.562 \\
& EchoMimicV3 & 1.3B & 3.50 & \underline{46.797} & 347.066 & 2.397 & 12.455 & 2.425 \\
& AniTalker & 0.1B & \underline{2.66} & 95.210 & 627.765 & \textbf{6.145} & \underline{8.610} & \underline{1.538} \\
& Ditto & 0.1B & \underline{2.66} & \textbf{28.503} & \underline{329.314} & 4.412 & 9.767 & 2.004 \\
\rowcolor{first}& Ours & 0.6B & \textbf{1.50} & 55.682 & \textbf{316.275} & \underline{5.866} & \textbf{8.459} & \textbf{0.872} \\
\bottomrule
\end{tabular}
}
\caption{Quantitative comparison with several state-of-the-art methods methods on HDTF and MEAD datasets. We use bold to indicate the best score and underline to represent the second-best score. “↑” indicates better performance with higher values, while “↓” indicates better performance with lower values. \colorbox{first}{EARTalking} achieved the best overall performance.}
\label{tab: Overall Comparison}
\vspace{-10pt}
\end{table*}

\subsection{Overall Comparison}

\textbf{Baseline.}
We compared our method against typical outstanding works across various technical routes. Among these, AniPortrait\cite{aniportrait}, EchoMimic\cite{chen2025echomimic} are classic Diffusion-based works. EchoMimicV3\cite{meng2025echomimicv3} is a Diffusion-based approach, leveraging a DiT backbone combined with 3D VAE for spatiotemporal compression. In contrast, methods like AniTalker\cite{anitalker}, Ditto\cite{li2025ditto}, and ARTalk\cite{chu2025artalk} follow a non-end-to-end technical route. This involves predicting an intermediate motion representation from the input audio, which is then used to render the final video. Notably, ARTalk utilizes a causal autoregressive scheme for this process.

As shown in Table \ref{tab: Overall Comparison}, previous methods struggle to achieve a balanced set of performance metrics due to their specialized designs. 
For instance, Ditto's strict motion constraint yields a favorable FID but causes poor lip synchronization and an extremely low E-FID (failing to preserve expression). Conversely, Anitalker achieves a high Sync-C by focusing on coarse mouth movements, but this approach leads to severe visual distortions, unnatural expressions, and a very high (poor) FVD score.
In contrast, our model demonstrates comprehensive and superior performance. Our novel in-context streaming condition approach yields the outstanding Syn-C and Syn-D metrics for outstanding lip synchronization, and  the best E-FID results for the exceptional ability to maintain identity and expression consistency. Furthermore, our reference feature anchoring and kv-cache preservation ensure high visual quality and temporal stability. This results in the best (low) FVD score, demonstrating superior video quality and temporal coherence over previous methods.

\begin{figure}[t] 
    \centering
    \includegraphics[width=\columnwidth]{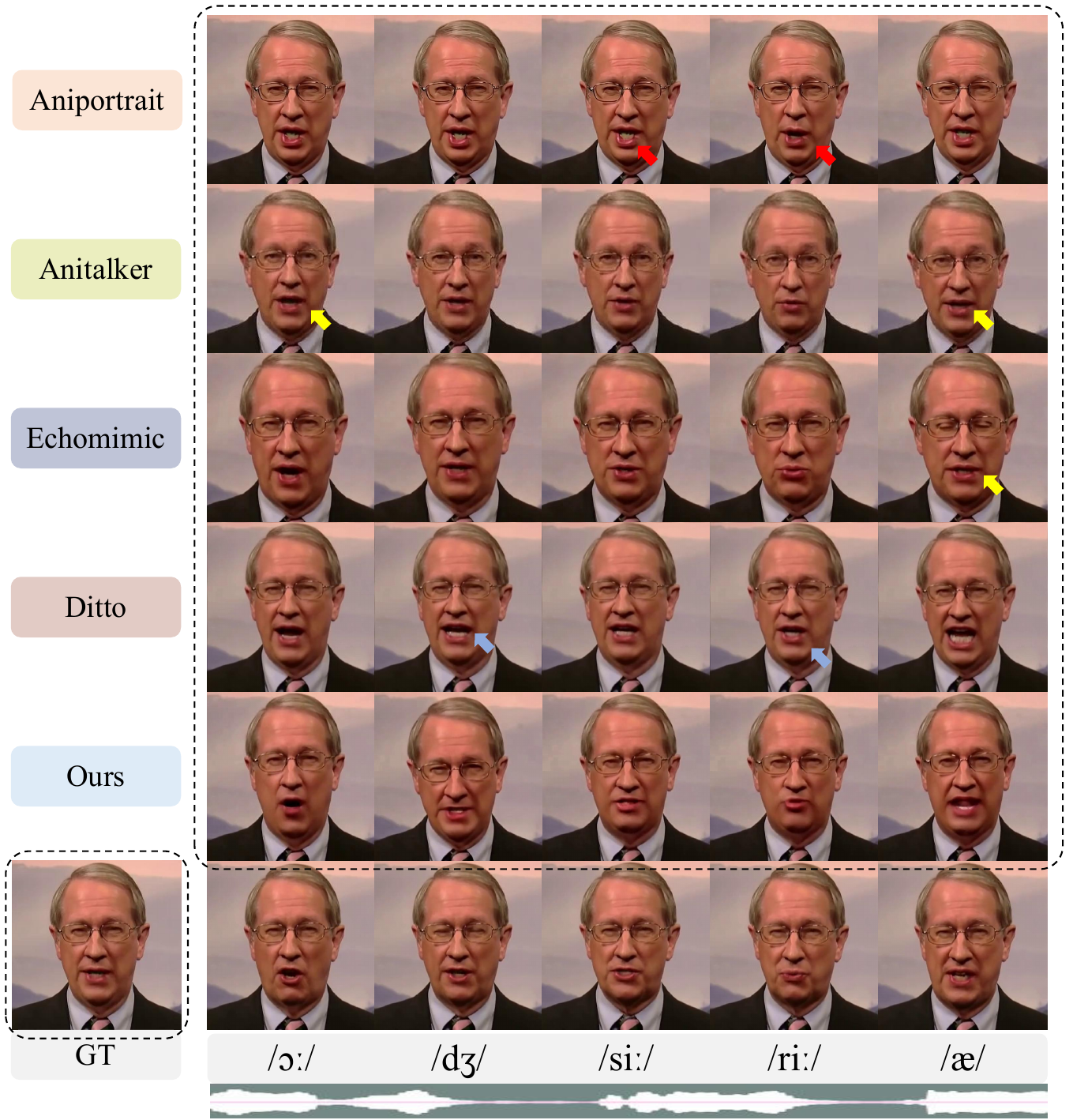}
    \caption{Qualitative comparison on the HDTF against leading methods with diverse technical routes. The red arrow points to lip distortion, the yellow arrow points to insufficient mouth opening, and the blue arrow points to insufficient lip closure.}
    \label{tab:quantitative_comparison}
    \vspace{-12pt}
\end{figure}

\subsection{Case Study}

Figure \ref{tab:quantitative_comparison} presents a qualitative comparison of video results from different audio-driven talking head methods. The bottom of the figure displays the corresponding audio waveform and phoneme sequence for each frame. The last row shows the ground truth video frames (right) and the reference portrait image (left) used to drive each method.

The results indicate that Aniportrait exhibit poor lip-audio synchronization throughout the sequence, struggling particularly with inadequate mouth closure for closed-mouth phonemes. Anitalker shows misalignment at the beginning and end of the video and performs poorly on open-mouth sounds. Echomimic is constrained by its need to crop the reference image, which hinders its ability to remain faithful to the original video's composition. Despite being the outstanding causal autoregressive method, ARTalk's non-end-to-end design poorly interprets speech, causing minimal lip motion. Its multi-stage complexity also necessitates background/pose changes, reducing fidelity to the reference image. In contrast, our method demonstrates superior performance by maintaining both precise lip synchronization aligned with the audio and natural, expressive facial movements compared to these other approaches.

\subsection{Ablation Study}

\textbf{Frame Condition In-Context Strategies.}
As an end-to-end causal model that supports streaming generation from streaming input conditions, we conducted an ablation study to determine the optimal "frame condition in-context control" strategy, comparing different attention mask and kv-cache designs (results in Table \ref{tab:conditioning_strategy_ablation_study}).

\begin{table}[t]
\centering
\resizebox{\columnwidth}{!}{
\tiny 
\begin{tabular}{c|ccc}
\toprule
\textbf{Method} & \textbf{FID} ($\downarrow$) & \textbf{FVD} ($\downarrow$) & \textbf{Sync-C} ($\uparrow$) \\
\midrule
Ours & 18.981 & \textbf{363.909} & \textbf{5.707} \\
w/o Audio-Cache & 19.136 & 389.641 & 5.329  \\
Uni-Att & \textbf{18.876} & 410.619 & 5.336 \\
\bottomrule
\end{tabular}
}
\caption{Comparison of different audio conditioning strategies. ours indicates that the kv-cache stores both past visual and audio representations. w/o audio-cache refers to not storing past audio representations in the kv-cache. uni-att denotes that only uni-directional attention from visual tokens to audio tokens is retained.} 
\label{tab:conditioning_strategy_ablation_study}
\vspace{-12pt}
\end{table}

Experiments demonstrate that our model, by maintaining an audio cache, can integrate current-frame control conditions with historical audio and visual information to capture better lip movements. When the audio cache is removed, the Syn-C metric shows a significant drop, while FID and FVD still perform excellently. This demonstrates the design's effectiveness in enhancing the model's adherence to conditional instructions, validating it as a feasible and efficient solution for streaming autoregressive generation.

Furthermore, we validated the attention mechanism itself. Our full model design permits audio-to-video attention (i.e., audio can attend to all past frames, including the reference image). We tested a uni-att variant, which retained only unidirectional video-to-audio attention. This uni-att configuration resulted in a degradation of both Sync-C and FVD scores, validating the necessity and effectiveness of bidirectional attention between modalities in this autoregressive paradigm.

\begin{figure}[t] 
    \centering
    \includegraphics[width=\columnwidth]{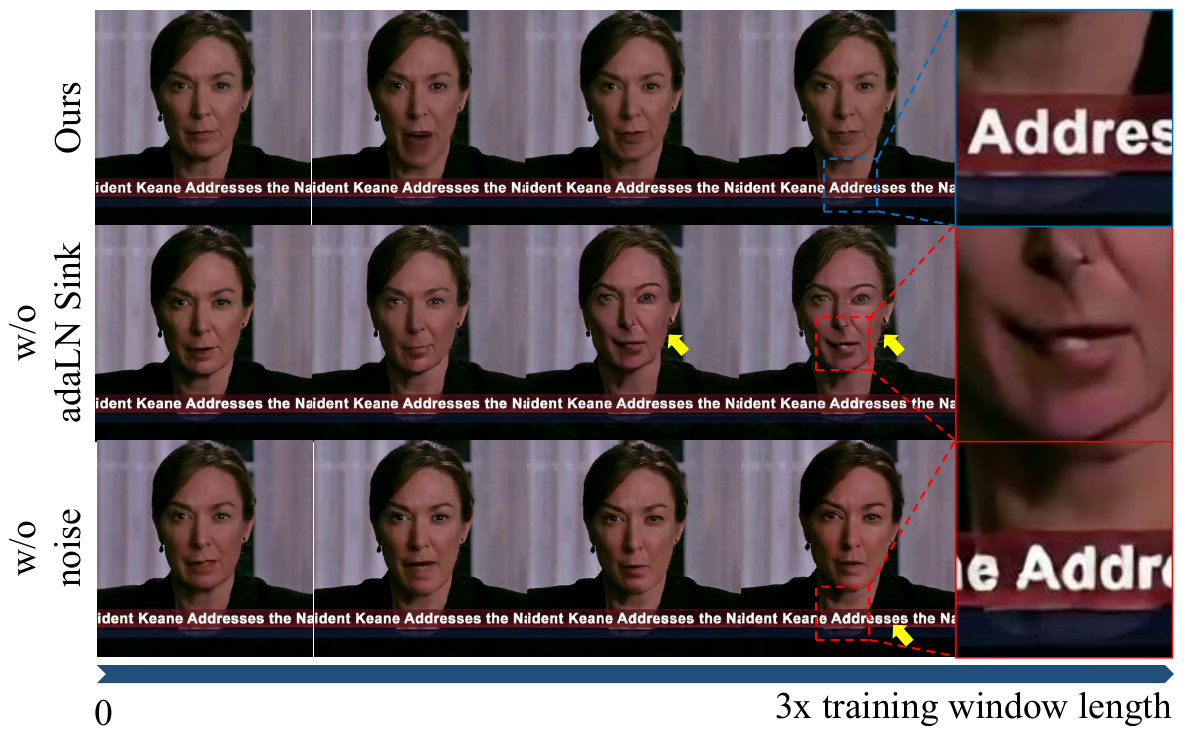}
    \caption{Ablation study for inference at 3x the training window length. w/o adaLN sink: The adaLN sink on the reference image is removed during both training and inference. w/o Noise: Noise perturbation to the input video is disabled during training.}
    \label{fig:case_study_length}
    \vspace{-5pt}
\end{figure}

\begin{table}[t]
\centering
\resizebox{\columnwidth}{!}{
\tiny 
\begin{tabular}{c|ccc}
\toprule
\textbf{Method} & \textbf{FID} ($\downarrow$) & \textbf{FVD} ($\downarrow$) & \textbf{Sync-C} ($\uparrow$) \\
\midrule
Ours & \textbf{18.981} & \textbf{363.909} & \textbf{5.707}  \\
w/o Noise & 20.643 & 401.354 & 4.277   \\
w/o adaLN & 42.298 & 489.197  & 4.067 \\
\bottomrule
\end{tabular}
}
\caption{Ablation study results for noise perturbation and adaLN sink mentioned in \ref{3.3 Sink Frame Window Attention Framework}. w/o Noise denotes the removal of noise perturbation during training, and w/o adaLN denotes the removal of the adaLN sink design during training and inference.} 
\label{tab:ariable-length_ablation_study}
\vspace{-12pt}
\end{table}

\noindent\textbf{Autoregressive Robustness Ablation Study.}
We conducted an ablation study to validate our method's variable-length generation and its ability to mitigate autoregressive error. We compared our full method (using both noise perturbation and adaLN sink anchoring) against two variants: one without noise and one without adaLN sink anchoring. The results in Table \ref{tab:ariable-length_ablation_study} show that removing either component degrades performance. Qualitative examples in Figure \ref{fig:case_study_length} further illustrate this: without noise perturbation (row 2), autoregressive errors accumulate, causing fonts in later frames to become noticeably distorted. Similarly, removing the adaLN sink anchoring (row 3) not only causes this same font distortion but also impairs motion control, leading to severe facial distortion.

\begin{figure}[t] 
    \centering
    \includegraphics[width=\columnwidth]{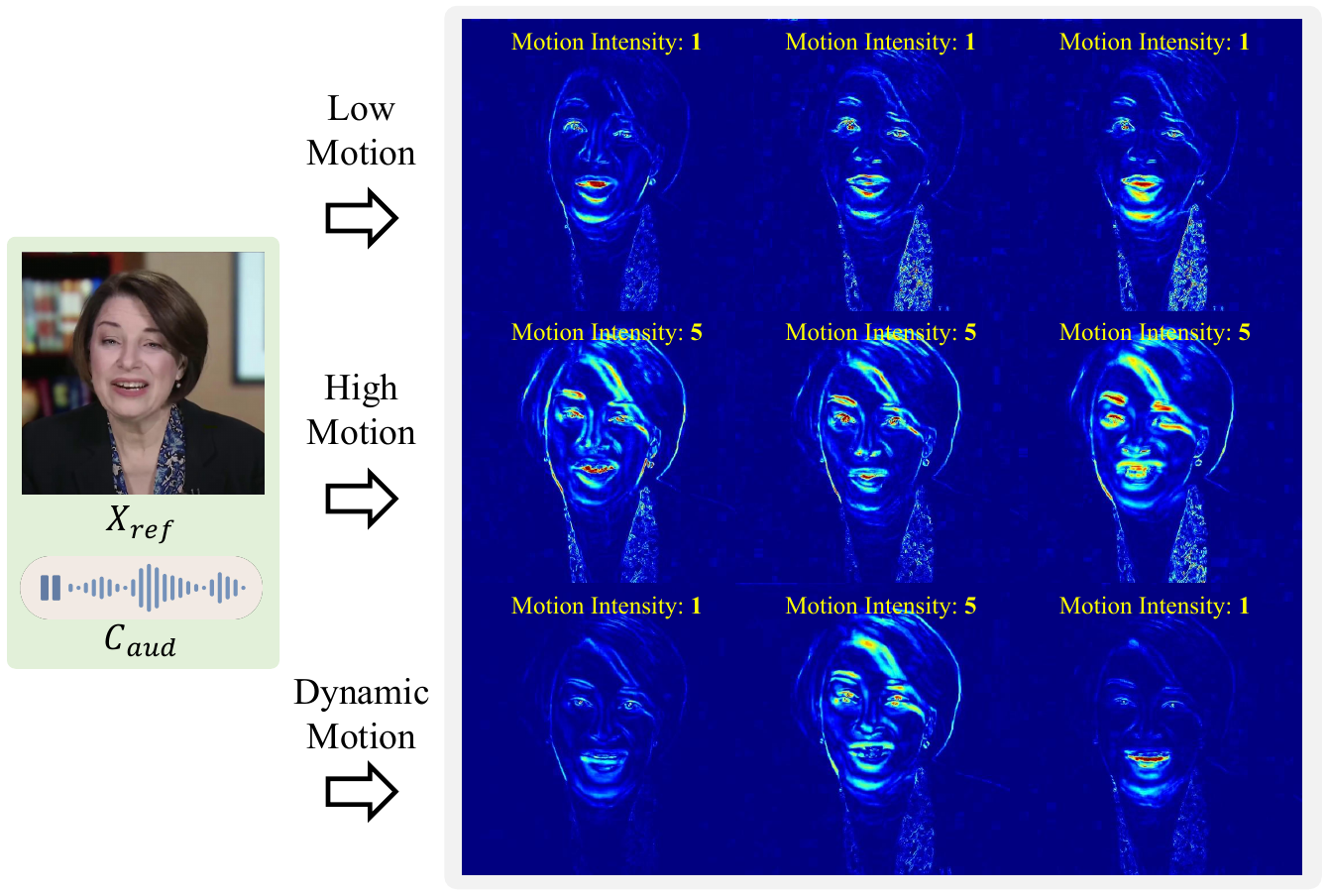}
    \caption{The difference heatmap of the correlation between the generated character's motion and the injected motion intensity.}
    \label{case_study}
    \label{fig:motion_case}
    \vspace{-12pt}
\end{figure}

\noindent\textbf{Frame-Level Condition Control Study.}
To evaluate the effectiveness of contextual learning and auto-regressive generation for streaming, frame-level control, we conducted an experiment using varied motion intensity conditions while holding the driving audio and reference image constant. As illustrated in Figure \ref{fig:motion_case}, when a uniformly low motion intensity is applied to the entire sequence, the generated character exhibits minimal head movement, with motion primarily localized to the labial (lip) and ocular (eye) regions. Conversely, applying a uniformly high motion intensity significantly amplifies the magnitude of movement in both facial expressions and head pose. Finally, when we injected a different motion intensity value for each frame during the auto-regressive process, the resulting motion demonstrated a high fidelity to the frame-specific control conditions, without being adversely influenced by adjacent temporal frames. This experiment validates that our method of contextual condition injection is effective for multi-conditional control and successfully achieves fine-grained temporal granularity. This control scheme remains effective and highly scalable, even when incorporating other types of control conditions.



\section{Conclusion}
In this paper, we introduce EARTalking, the first end-to-end, GPT-style autoregressive model for talking head generation. EARTalking presents a novel, high-performance, and highly scalable solution that facilitates multi-conditional, frame-by-frame streaming control. This approach uniquely allows users to adjust the generated output at any arbitrary frame, bypassing the fixed-window latency of prior methods.

\noindent\textbf{Limitations and Future Work}
Although EARTalking achieves low computational cost and a small memory footprint, it still struggles to reach real-time targets due to its masked autoregressive inference. The experiments validate the effectiveness of the autoregressive architecture for streaming, multi-conditional control (primarily audio and motion). Future work could therefore focus on integrating additional control conditions into the streaming framework to enhance its practical utility.


{
    \small
    \bibliographystyle{ieeenat_fullname}
    \bibliography{main}

@String(CVPR= {IEEE Conf. Comput. Vis. Pattern Recog.})

@String(ICCV= {Int. Conf. Comput. Vis.})

@String(ICLR = {Int. Conf. Learn. Represent.})

@String(AAAI = {AAAI})

@String(CVPR  = {CVPR})

@String(ICCV  = {ICCV})

@String(ICLR  = {ICLR})

@article{thdsurvey,
  title={Human-computer interaction system: A survey of talking-head generation},
  author={Zhen, Rui and Song, Wenchao and He, Qiang and Cao, Juan and Shi, Lei and Luo, Jia},
  journal={Electronics},
  volume={12},
  number={1},
  pages={218},
  year={2023},
  publisher={MDPI}
}

@article{xu2024hallo,
  title={Hallo: Hierarchical audio-driven visual synthesis for portrait image animation},
  author={Xu, Mingwang and Li, Hui and Su, Qingkun and Shang, Hanlin and Zhang, Liwei and Liu, Ce and Wang, Jingdong and Yao, Yao and Zhu, Siyu},
  journal={arXiv preprint arXiv:2406.08801},
  year={2024}
}

@inproceedings{tian2024emo,
  title={Emo: Emote portrait alive generating expressive portrait videos with audio2video diffusion model under weak conditions},
  author={Tian, Linrui and Wang, Qi and Zhang, Bang and Bo, Liefeng},
  booktitle={European Conference on Computer Vision},
  pages={244--260},
  year={2024},
  organization={Springer}
}

@inproceedings{chen2025echomimic,
  title={Echomimic: Lifelike audio-driven portrait animations through editable landmark conditions},
  author={Chen, Zhiyuan and Cao, Jiajiong and Chen, Zhiquan and Li, Yuming and Ma, Chenguang},
  booktitle={Proceedings of the AAAI Conference on Artificial Intelligence},
  volume={39},
  number={3},
  pages={2403--2410},
  year={2025}
}

@inproceedings{tan2024flowvqtalker,
  title={Flowvqtalker: High-quality emotional talking face generation through normalizing flow and quantization},
  author={Tan, Shuai and Ji, Bin and Pan, Ye},
  booktitle={Proceedings of the IEEE/CVF Conference on Computer Vision and Pattern Recognition},
  pages={26317--26327},
  year={2024}
}

@inproceedings{jang2024faces,
  title={Faces that speak: Jointly synthesising talking face and speech from text},
  author={Jang, Youngjoon and Kim, Ji-Hoon and Ahn, Junseok and Kwak, Doyeop and Yang, Hong-Sun and Ju, Yoon-Cheol and Kim, Il-Hwan and Kim, Byeong-Yeol and Chung, Joon Son},
  booktitle={Proceedings of the IEEE/CVF Conference on Computer Vision and Pattern Recognition},
  pages={8818--8828},
  year={2024}
}

@inproceedings{peng2024synctalk,
  title={Synctalk: The devil is in the synchronization for talking head synthesis},
  author={Peng, Ziqiao and Hu, Wentao and Shi, Yue and Zhu, Xiangyu and Zhang, Xiaomei and Zhao, Hao and He, Jun and Liu, Hongyan and Fan, Zhaoxin},
  booktitle={Proceedings of the IEEE/CVF Conference on Computer Vision and Pattern Recognition},
  pages={666--676},
  year={2024}
}

@article{CPE,
  author={Mukai Li and
                  Shansan Gong and
                  Jiangtao Feng and
                  Yiheng Xu and
                  Jun Zhang and
                  Zhiyong Wu and
                  Lingpeng Kong},
  title={In-Context Learning with Many Demonstration Examples},
  journal={CoRR},
  volume={abs/2302.04931},
  year={2023},
  url={https://doi.org/10.48550/arXiv.2302.04931},
  doi={10.48550/ARXIV.2302.04931},
  eprinttype={arXiv},
  eprint={2302.04931},
  bibsource={dblp computer science bibliography, https://dblp.org}
}

@inproceedings{EmotiveTalk,
  author       = {Haotian Wang and
                  Yuzhe Weng and
                  Yueyan Li and
                  Zilu Guo and
                  Jun Du and
                  Shutong Niu and
                  Jiefeng Ma and
                  Shan He and
                  Xiaoyan Wu and
                  Qiming Hu and
                  Bing Yin and
                  Cong Liu and
                  Qingfeng Liu},
  title        = {EmotiveTalk: Expressive Talking Head Generation through Audio Information
                  Decoupling and Emotional Video Diffusion},
  booktitle    = {{IEEE/CVF} Conference on Computer Vision and Pattern Recognition,
                  {CVPR} 2025, Nashville, TN, USA, June 11-15, 2025},
  pages        = {26212--26221},
  publisher    = {Computer Vision Foundation / {IEEE}},
  year         = {2025},
  url          = {https://openaccess.thecvf.com/content/CVPR2025/html/Wang\_EmotiveTalk\_Expressive\_Talking\_Head\_Generation\_through\_Audio\_Information\_Decoupling\_and\_CVPR\_2025\_paper.html},
  doi          = {10.1109/CVPR52734.2025.02441},
  bibsource    = {dblp computer science bibliography, https://dblp.org}
}

@InProceedings{Gan_2023_ICCV,
        author    = {Gan, Yuan and Yang, Zongxin and Yue, Xihang and Sun, Lingyun and Yang, Yi},
        title     = {Efficient Emotional Adaptation for Audio-Driven Talking-Head Generation},
        booktitle = {Proceedings of the IEEE/CVF International Conference on Computer Vision (ICCV)},
        month     = {October},
        year      = {2023},
        pages     = {22634-22645}
    }

@article{PC-Talk,
  author       = {Baiqin Wang and
                  Xiangyu Zhu and
                  Fan Shen and
                  Hao Xu and
                  Zhen Lei},
  title        = {PC-Talk: Precise Facial Animation Control for Audio-Driven Talking
                  Face Generation},
  journal      = {CoRR},
  volume       = {abs/2503.14295},
  year         = {2025},
  url          = {https://doi.org/10.48550/arXiv.2503.14295},
  doi          = {10.48550/ARXIV.2503.14295},
  eprinttype    = {arXiv},
  eprint       = {2503.14295},
  bibsource    = {dblp computer science bibliography, https://dblp.org}
}

@article{SVD,
  title={Stable video diffusion: Scaling latent video diffusion models to large datasets},
  author={Blattmann, Andreas and Dockhorn, Tim and Kulal, Sumith and Mendelevitch, Daniel and Kilian, Maciej and Lorenz, Dominik and Levi, Yam and English, Zion and Voleti, Vikram and Letts, Adam and others},
  journal={arXiv preprint arXiv:2311.15127},
  year={2023}
}

@article{ViT,
  title={An image is worth 16x16 words: Transformers for image recognition at scale},
  author={Dosovitskiy, Alexey},
  journal={arXiv preprint arXiv:2010.11929},
  year={2020}
}

@inproceedings{DiT,
  title={Scalable diffusion models with transformers},
  author={Peebles, William and Xie, Saining},
  booktitle={Proceedings of the IEEE/CVF international conference on computer vision},
  pages={4195--4205},
  year={2023}
}

@article{kong2024hunyuanvideo,
  title={Hunyuanvideo: A systematic framework for large video generative models},
  author={Kong, Weijie and Tian, Qi and Zhang, Zijian and Min, Rox and Dai, Zuozhuo and Zhou, Jin and Xiong, Jiangfeng and Li, Xin and Wu, Bo and Zhang, Jianwei and others},
  journal={arXiv preprint arXiv:2412.03603},
  year={2024}
}

@article{wan2025,
      title={Wan: Open and Advanced Large-Scale Video Generative Models}, 
      author={Team Wan and Ang Wang and Baole Ai and Bin Wen and Chaojie Mao and Chen-Wei Xie and Di Chen and Feiwu Yu and Haiming Zhao and Jianxiao Yang and Jianyuan Zeng and Jiayu Wang and Jingfeng Zhang and Jingren Zhou and Jinkai Wang and Jixuan Chen and Kai Zhu and Kang Zhao and Keyu Yan and Lianghua Huang and Mengyang Feng and Ningyi Zhang and Pandeng Li and Pingyu Wu and Ruihang Chu and Ruili Feng and Shiwei Zhang and Siyang Sun and Tao Fang and Tianxing Wang and Tianyi Gui and Tingyu Weng and Tong Shen and Wei Lin and Wei Wang and Wei Wang and Wenmeng Zhou and Wente Wang and Wenting Shen and Wenyuan Yu and Xianzhong Shi and Xiaoming Huang and Xin Xu and Yan Kou and Yangyu Lv and Yifei Li and Yijing Liu and Yiming Wang and Yingya Zhang and Yitong Huang and Yong Li and You Wu and Yu Liu and Yulin Pan and Yun Zheng and Yuntao Hong and Yupeng Shi and Yutong Feng and Zeyinzi Jiang and Zhen Han and Zhi-Fan Wu and Ziyu Liu},
      journal = {arXiv preprint arXiv:2503.20314},
      year={2025}
}

@article{wang2024emu3,
  title={Emu3: Next-token prediction is all you need},
  author={Wang, Xinlong and Zhang, Xiaosong and Luo, Zhengxiong and Sun, Quan and Cui, Yufeng and Wang, Jinsheng and Zhang, Fan and Wang, Yueze and Li, Zhen and Yu, Qiying and others},
  journal={arXiv preprint arXiv:2409.18869},
  year={2024}
}

@InProceedings{deng2024nova,
        author    = {Deng, Haoge and Pan, Ting and Diao, Haiwen and Luo, Zhengxiong and Cui, Yufeng and Lu, Huchuan and Shan, Shiguang and Qi, Yonggang and Wang, Xinlong},
        title     = {Autoregressive Video Generation without Vector Quantization},
        booktitle = {The International Conference on Learning Representations (ICLR)},
        year      = {2025},
    }

@article{yu2025videomar,
author = {Yu, Hu and Gong, Biao and Yuan, Hangjie and Zheng, DanDan and Chai, Weilong and Chen, Jingdong and Zheng, Kecheng and Zhao, Feng},
year = {2025},
month = {06},
pages = {},
title = {VideoMAR: Autoregressive Video Generatio with Continuous Tokens},
doi = {10.48550/arXiv.2506.14168}
}

@inproceedings{StyleTalk,
author = {Ma, Yifeng and Wang, Suzhen and Hu, Zhipeng and Fan, Changjie and Lv, Tangjie and Ding, Yu and Deng, Zhidong and Yu, Xin},
title = {StyleTalk: one-shot talking head generation with controllable speaking styles},
year = {2023},
isbn = {978-1-57735-880-0},
publisher = {AAAI Press},
url = {https://doi.org/10.1609/aaai.v37i2.25280},
doi = {10.1609/aaai.v37i2.25280},
booktitle = {Proceedings of the Thirty-Seventh AAAI Conference on Artificial Intelligence and Thirty-Fifth Conference on Innovative Applications of Artificial Intelligence and Thirteenth Symposium on Educational Advances in Artificial Intelligence},
articleno = {211},
numpages = {9},
series = {AAAI'23/IAAI'23/EAAI'23}
}

@article{ma2023dreamtalk,
  title={DreamTalk: When Expressive Talking Head Generation Meets Diffusion Probabilistic Models},
  author={Ma, Yifeng and Zhang, Shiwei and Wang, Jiayu and Wang, Xiang and Zhang, Yingya and Deng, Zhidong},
  journal={arXiv preprint arXiv:2312.09767},
  year={2023}
}

@article{cheng2024dawn,
  title={DAWN: Dynamic Frame Avatar with Non-autoregressive Diffusion Framework for Talking Head Video Generation},
  author={Cheng, Hanbo and Lin, Limin and Liu, Chenyu and Xia, Pengcheng and Hu, Pengfei and Ma, Jiefeng and Du, Jun and Pan, Jia},
  journal={arXiv preprint arXiv:2410.13726},
  year={2024}
}

@misc{
    chu2025artalk,
    title={ARTalk: Speech-Driven 3D Head Animation via Autoregressive Model}, 
    author={Xuangeng Chu and Nabarun Goswami and Ziteng Cui and Hanqin Wang and Tatsuya Harada},
    year={2025},
    eprint={2502.20323},
    archivePrefix={arXiv},
    primaryClass={cs.CV},
    url={https://arxiv.org/abs/2502.20323}, 
}

@inproceedings{sungbin24_interspeech,
  title     = {MultiTalk: Enhancing 3D Talking Head Generation Across Languages with Multilingual Video Dataset},
  author    = {Kim Sung-Bin and Lee Chae-Yeon and Gihun Son and Oh Hyun-Bin and Janghoon Ju and Suekyeong Nam and Tae-Hyun Oh},
  year      = {2024},
  booktitle = {Interspeech 2024},
  pages     = {1380--1384},
  doi       = {10.21437/Interspeech.2024-1794},
  issn      = {2958-1796},
}

@inproceedings{nocentini2024emovocaspeechdrivenemotional3d,
title={EmoVOCA: Speech-Driven Emotional 3D Talking Heads}, 
author={Federico Nocentini and Claudio Ferrari and Stefano Berretti},
booktitle = {Proceedings of the IEEE/CVF Winter Conference on Applications of Computer Vision (WACV)},
year = {2025},
}

@inproceedings{ma2025exploring,
  title={Exploring Timeline Control for Facial Motion Generation},
  author={Ma, Yifeng and Qi, Jinwei and Ji, Chaonan and Zhang, Peng and Zhang, Bang and Deng, Zhidong and Bo, Liefeng},
  booktitle={Proceedings of the Computer Vision and Pattern Recognition Conference},
  pages={1940--1950},
  year={2025}
}

@inproceedings{li2025portrait,
  title={IM-Portrait: Learning 3D-aware Video Diffusion for Photorealistic Talking Heads from Monocular VideosC},
  author={Li, Yuan and Bai, Ziqian and Tan, Feitong and Cui, Zhaopeng and Fanello, Sean and Zhang, Yinda},
  booktitle={Proceedings of the Computer Vision and Pattern Recognition Conference},
  pages={21107--21116},
  year={2025}
}

@inproceedings{chen2025taoavatar,
  title={TaoAvatar: Real-Time Lifelike Full-Body Talking Avatars for Augmented Reality via 3D Gaussian Splatting},
  author={Chen, Jianchuan and Hu, Jingchuan and Wang, Gaige and Jiang, Zhonghua and Zhou, Tiansong and Chen, Zhiwen and Lv, Chengfei},
  booktitle={Proceedings of the Computer Vision and Pattern Recognition Conference},
  pages={10723--10734},
  year={2025}
}

@misc{hu2025HunyuanVideo-Avatar,
      title={HunyuanVideo-Avatar: High-Fidelity Audio-Driven Human Animation for Multiple Characters}, 
      author={Yi Chen and Sen Liang and Zixiang Zhou and Ziyao Huang and Yifeng Ma and Junshu Tang and Qin Lin and Yuan Zhou and Qinglin Lu},
      year={2025},
      eprint={2505.20156},
      archivePrefix={arXiv},
      primaryClass={cs.CV},
      url={https://arxiv.org/pdf/2505.20156}, 
}

@article{fid,
  title={Gans trained by a two time-scale update rule converge to a local nash equilibrium},
  author={Heusel, Martin and Ramsauer, Hubert and Unterthiner, Thomas and Nessler, Bernhard and Hochreiter, Sepp},
  journal={Advances in neural information processing systems},
  volume={30},
  year={2017}
}

@article{fvd,
  title={FVD: A new metric for video generation},
  author={Unterthiner, Thomas and van Steenkiste, Sjoerd and Kurach, Karol and Marinier, Rapha{\"e}l and Michalski, Marcin and Gelly, Sylvain},
  year={2019}
}

@misc{stylegan_v,
    title={StyleGAN-V: A Continuous Video Generator with the Price, Image Quality and Perks of StyleGAN2},
    author={Ivan Skorokhodov and Sergey Tulyakov and Mohamed Elhoseiny},
    journal={arXiv preprint arXiv:2112.14683},
    year={2021}
}

@article{mar,
  title={Autoregressive image generation without vector quantization},
  author={Li, Tianhong and Tian, Yonglong and Li, He and Deng, Mingyang and He, Kaiming},
  journal={Advances in Neural Information Processing Systems},
  volume={37},
  pages={56424--56445},
  year={2024}
}

@inproceedings{Transformer,
 author = {Vaswani, Ashish and Shazeer, Noam and Parmar, Niki and Uszkoreit, Jakob and Jones, Llion and Gomez, Aidan N and Kaiser, \L ukasz and Polosukhin, Illia},
 booktitle = {Advances in Neural Information Processing Systems},
 editor = {I. Guyon and U. Von Luxburg and S. Bengio and H. Wallach and R. Fergus and S. Vishwanathan and R. Garnett},
 pages = {},
 publisher = {Curran Associates, Inc.},
 title = {Attention is All you Need},
 url = {https://proceedings.neurips.cc/paper_files/paper/2017/file/3f5ee243547dee91fbd053c1c4a845aa-Paper.pdf},
 volume = {30},
 year = {2017}
}

@article{xie2024show,
  title={Show-o: One single transformer to unify multimodal understanding and generation},
  author={Xie, Jinheng and Mao, Weijia and Bai, Zechen and Zhang, David Junhao and Wang, Weihao and Lin, Kevin Qinghong and Gu, Yuchao and Chen, Zhijie and Yang, Zhenheng and Shou, Mike Zheng},
  journal={arXiv preprint arXiv:2408.12528},
  year={2024}
}

@article{chen2025blip3,
  title={Blip3-o: A family of fully open unified multimodal models-architecture, training and dataset},
  author={Chen, Jiuhai and Xu, Zhiyang and Pan, Xichen and Hu, Yushi and Qin, Can and Goldstein, Tom and Huang, Lifu and Zhou, Tianyi and Xie, Saining and Savarese, Silvio and others},
  journal={arXiv preprint arXiv:2505.09568},
  year={2025}
}

@inproceedings{xie2025muse,
  title={Muse-vl: Modeling unified vlm through semantic discrete encoding},
  author={Xie, Rongchang and Du, Chen and Song, Ping and Liu, Chang},
  booktitle={Proceedings of the IEEE/CVF International Conference on Computer Vision},
  pages={24135--24146},
  year={2025}
}

@article{chen2022wavlm,
  title={Wavlm: Large-scale self-supervised pre-training for full stack speech processing},
  author={Chen, Sanyuan and Wang, Chengyi and Chen, Zhengyang and Wu, Yu and Liu, Shujie and Chen, Zhuo and Li, Jinyu and Kanda, Naoyuki and Yoshioka, Takuya and Xiao, Xiong and others},
  journal={IEEE Journal of Selected Topics in Signal Processing},
  volume={16},
  number={6},
  pages={1505--1518},
  year={2022},
  publisher={IEEE}
}

@inproceedings{hdtf,
  title={Flow-guided one-shot talking face generation with a high-resolution audio-visual dataset},
  author={Zhang, Zhimeng and Li, Lincheng and Ding, Yu and Fan, Changjie},
  booktitle={Proceedings of the IEEE/CVF Conference on Computer Vision and Pattern Recognition},
  pages={3661--3670},
  year={2021}
}

@inproceedings{mead,
  title={Mead: A large-scale audio-visual dataset for emotional talking-face generation},
  author={Wang, Kaisiyuan and Wu, Qianyi and Song, Linsen and Yang, Zhuoqian and Wu, Wayne and Qian, Chen and He, Ran and Qiao, Yu and Loy, Chen Change},
  booktitle={European Conference on Computer Vision},
  pages={700--717},
  year={2020},
  organization={Springer}
}

@inproceedings{hallo3,
  title={Hallo3: Highly dynamic and realistic portrait image animation with video diffusion transformer},
  author={Cui, Jiahao and Li, Hui and Zhan, Yun and Shang, Hanlin and Cheng, Kaihui and Ma, Yuqi and Mu, Shan and Zhou, Hang and Wang, Jingdong and Zhu, Siyu},
  booktitle={Proceedings of the Computer Vision and Pattern Recognition Conference},
  pages={21086--21095},
  year={2025}
}

@article{aniportrait,
  title={Aniportrait: Audio-driven synthesis of photorealistic portrait animation},
  author={Wei, Huawei and Yang, Zejun and Wang, Zhisheng},
  journal={arXiv preprint arXiv:2403.17694},
  year={2024}
}

@misc{meng2025echomimicv3,
  title={EchoMimicV3: 1.3B Parameters are All You Need for Unified Multi-Modal and Multi-Task Human Animation},
  author={Rang Meng and Yan Wang and Weipeng Wu and Ruobing Zheng and Yuming Li and Chenguang Ma},
  year={2025},
  eprint={2507.03905},
  archivePrefix={arXiv}
}

@article{anitalker,
  title={AniTalker: Animate Vivid and Diverse Talking Faces through Identity-Decoupled Facial Motion Encoding},
  author={Liu, Tao and Chen, Feilong and Fan, Shuai and Du, Chenpeng and Chen, Qi and Chen, Xie and Yu, Kai},
  journal={arXiv preprint arXiv:2405.03121},
  year={2024}
}

@inproceedings{li2025ditto,
  title={Ditto: Motion-space diffusion for controllable realtime talking head synthesis},
  author={Li, Tianqi and Zheng, Ruobing and Yang, Minghui and Chen, Jingdong and Yang, Ming},
  booktitle={Proceedings of the 33rd ACM International Conference on Multimedia},
  pages={9704--9713},
  year={2025}
}
}



\end{document}